\documentclass[11pt]{article}
\usepackage{blindtext}
\usepackage{algorithmicx}
\usepackage{algorithm}% http://ctan.org/pkg/algorithms
\usepackage{algpseudocode}% http://ctan.org/pkg/algorithmicx
\usepackage{graphicx}
\usepackage{rotating}
\usepackage{url}
\usepackage{amsmath}
\usepackage{multirow}
\usepackage{graphicx}
\usepackage{amsmath}
\usepackage{multirow}

\PassOptionsToPackage{sectionbib,square,comma,numbers,sort&compress,super}{natbib} 
\usepackage[numbers]{natbib}
\usepackage{hyperref}       % hyperlinks
\usepackage{url}         \usepackage[utf8]{inputenc} % allow utf-8 input
\usepackage[T1]{fontenc}    % use 8-bit T1 fonts
\usepackage{hyperref}       % hyperlinks
\usepackage{url}            % simple URL typesetting
\usepackage{booktabs}       % professional-quality tables
\usepackage{amsfonts}       % blackboard math symbols
\usepackage{nicefrac}       % compact symbols for 1/2, etc.
\usepackage{microtype}      % microtypography
\usepackage{lipsum}		%   % simple URL typesetting
\usepackage{amsfonts}       % blackboard math symbols
\usepackage{nicefrac}       % compact symbols for 1/2, etc.
\usepackage{microtype}      % microtypography
\usepackage{lipsum}		%
\usepackage{tabularx}
\usepackage{caption}
\usepackage{rotating}

\title{A Semantic Modular Framework for Events Topic Modeling in Social Media}

\author{Arya Hadizadeh Moghaddam, Saeedeh Momtazi\\
Computer Engineering Department\\
Amirkabir University of Technology, Tehran, Iran
}

\date{}

\begin{document}
\maketitle

\begin{abstract}
The advancement of social media contributes to the growing amount of content they share frequently. This framework provides a sophisticated place for people to report various real-life events. Detecting these events with the help of natural language processing has received researchers' attention, and various algorithms have been developed for this goal. In this paper, we propose a Semantic Modular Model (SMM) consisting of 5 different modules, namely Distributional Denoising Autoencoder, Incremental Clustering, Semantic Denoising, Defragmentation, and Ranking and Processing. The proposed model aims to (1) cluster various documents and ignore the documents that might not contribute to the identification of events, (2) identify more important and descriptive keywords. Compared to the state-of-the-art methods, the results show that the proposed model has a higher performance in identifying events with lower ranks and extracting keywords for more important events in three English Twitter datasets: FACup, SuperTuesday, and USElection. The proposed method outperformed the best reported results in the mean keyword-precision metric by 7.9\%.


\end{abstract}

\section{Introduction}
\label{sec:intro}

The extensive growth of social media in the past few years has caused people to join social media websites and contribute to the increasing amount of content on the Internet by sharing their daily activities. The huge amount of data shared on social media allows us to use this data for prediction in various tasks \citep{weiler2016evaluation}.  Many people share their day to day activities on social media. Such a collection of information might report a specific event \citep{weng2011event}; e.g., a player might score a goal in a football match, and people might report this event on their twitter account. Therefore, analyzing tweets in a specific time might identify this event. This makes event detection one of the popular tasks among researchers.  The event detection task can be more challenging than it looks, and it could be different from other social media analysis tasks \cite{peng2022reinforced}.

Event detection can be used in various fields, such as medicine \citep{jagannatha2016bidirectional}, emergency \citep{martinez2018twitter}, politics \citep{adedoyin2016rule}. The necessity of event detection in these fields comes from the fact that an important event is usually followed by a set of other events. For instance, a car accident is normally followed by traffic jams and casualties. Therefore, if the rescue team is being informed earlier and arrives on-time, they might prevent casualties. This indicates the importance of accurately detecting events within a suitable time interval.

%\subsection{Event Detection Approaches}
Event detection is normally performed using task-based or similarity-based approaches. Task-based methods first describe the problem that the system wants to solve. Then, the system gathers information as needed, and the classifier must be trained based on these data. Assume that we want to use a method in order to report a car accident \citep{ali2021traffic}. For this matter, data about the accident must be collected in a specific time interval, and then according to machine learning algorithms, the model must be trained. This allows us to have the ability to detect an event in a specific topic.
Similarity-based methods use a set of algorithms that are placed in a stream of data and can detect events by recognizing structures and similar patterns. They can detect various events using specific settings.

Task-based methods have a similar performance to text classification \citep{nugent2017comparison} and need supervised training, while run-time methods need to be efficient and be able to properly divide the events \citep{hasan2019real}. These methods are comprehensive, but they mostly need various parameters for different domains.

%\subsection{Event Detection Methods}

Event detection methods can be divided into these three categories:

\begin{enumerate}
	\item Document-based methods: In these methods, different documents (such as tweets) are clustered according to their similarity, such that structurally or semantically similar documents are grouped together in one cluster. Each cluster represents a different event. These methods mostly focus on the connection between the documents \citep{aiello2013sensing}.
	
	\item Feature-based methods: These methods are similar to topic modeling. They aim to output the words that represent a specific event. Some research studies on these methods focus on creating graphs in order to identify the keywords by considering their connection  \citep{saeed2019enhanced}.
	
	\item Classification-based methods: These methods need supervised training to assign each document in one of the predetermined classes based on their textual information. These methods are applicable in certain fields and cannot identify the event topics \citep{vongkusolkit2021situational}.
\end{enumerate}

The document-based methods and feature-based methods are used for topic modeling, and in this paper, our approach is to propose a new topic modeling approach for event detection. In topic modeling approaches, different events are ranked, and in every rank, there are  keywords representing the event. The other studies that are compared to our model are  based on topic modeling.

This research combines the document- and feature-based methods in order to make use of the advantages of both methods and minimize their weaknesses. The proposed method studies the connection between the documents, as well as the connection between the keywords. Furthermore, this research follows a module-based architecture adopted from \citep{xia2018random}. Our proposed architecture consists of 5 modules, and each step extracts useful information. The proposed method can also be used in a real-time scenario.

The structure of the paper is as follows: In Section \ref{sec:RelatedWorks} the related works are explained, and various structures are compared. Section \ref{sec:ProposedModel} introduces the proposed method along with its modules. Section \ref{sec:Results} presents experimental results with different metrics. Finally, the conclusion and future works are provided in Section \ref{sec:Conclusion}.

\section{Related Works}
\label{sec:RelatedWorks}

As mentioned in Section \ref{sec:intro}, three methods are mainly used for event detection, namely document-based, feature-based, and classification-based. In document-based methods, documents are placed in specific clusters according to their similarity to other documents through clustering. In feature-based methods, the keywords describing topics of various events are identified according to the stream of documents. In classification-based methods, a set of features are extracted from the document, and according to their labels, these features are classified into distinct classes.
In this section, we review the algorithms and models used in the literature. Considering the unsupervised behavior of document-based and feature-based methods which makes them more usable for various domains, we focus on these two groups.

\subsection{Document-based Methods}
In these methods, clustering is done based on the similarity among the extracted features from the texts, and each cluster can represent an event \citep{fedoryszak2019real}. \citet{petrovic2010streaming} proposed a model called Document-Pivot (Doc-p) Topic Detection. The process of clustering is accelerated in this method due to Local Sensitivity Hashing (LSH). Term Frequency - Inverse Document Frequency (TF-IDF) was used for extracting the vector of documents in order to review the co-occurrence of the document words. In this method, the new event can be better detected when the similarity between the new event and previous clusters is small. The basic clustering used in the Doc-p algorithms is Umass \citep{allan2000detections}.
	
One of the main problems of this method is that during the process of clustering, clusters are formed only based on the co-occurrence of words. However, there might be some frequent words inside the tweets that are not close in meaning and this algorithm fails to identify those.

\subsection{Feature-based Methods}
The Graph-Based Feature-Pivot (Gfeat-p) topic detection method was introduced by \citet{o2010tweetmotif}. Accordingly, each document is transformed into a graph, and then the clusters are computed with the help of the Structural Algorithm for Networks (SCAN) \citep{xu2007scan}. In order to detect events, this method focuses on the connections between the terms, as well as reviewing connected graphs.
In Soft Frequent Pattern Mining Algorithm (SFPM) which was introduced by \citet{aiello2013sensing}, frequent words are identified as well as the co-occurrence of the words in such a way that more than two terms are examined, and these recurring patterns will aid us in event detection and topic extraction. This model also uses a similarity-based method to avoid finding general and limited topics.

In the BNGram method which was introduced by \citet{aiello2013sensing}, n-grams are used for event detection instead of unigrams. The reason is that repetitive structures (such as retweets) might exist in the events.  DF-IDF is used for calculation, which is a helpful score to find frequent and similar patterns. In addition, Name Entity Recognition (NER) is used to demonstrate the importance of proper nouns in event detection \citep{phuvipadawat2010breaking}.
An exemplar-based method suggested by \citet{elbagoury2015exemplar} aims to search for tweets that are useful in describing an event or a certain topic. The idea behind this method is that each event can be represented with a tweet. Tweets with fewer overlaps with other topics and the most overlap with the related tweets of a topic are chosen as representatives.
Latent Dirichlet Allocation (LDA) is a widely used method introduced by \citet{blei2003latent}. Based on LDA, each document consists of a set of words and this is the only variable. The distribution of the topics is hidden from all documents and it needs to be calculated based on Bayesian connections.

Separable Non-negative Matrix Factorization (SNMF) method, which is introduced by \citet{prabandari2017comparative}, breaks the matrices in order to obtain the matrix for terms and topics and then the events are detected accordingly. In this method, original recovery, which uses algebraic manipulation, and KL recovery are utilized as a part of the algorithm.
In the method introduced by \citet{nur2015combination}, a combination of Singular Value Decomposition (SVD) and K-means is used, where the document matrices turn into factorized matrices, and then these matrices are clustered. Each cluster center is extracted and based on them the related keyword which describes the events are extracted.
\citet{saeed2019enhanced} introduced a method named Enhanced Heartbeat Graph (EHG) where the documents are transformed into a graph. Then, based on the recurring patterns of various word co-occurrences in time, these graphs are combined. The events are then detected based on different features, including divergence factor, trend probability, and topic centrality.

\citet{asgari2020topicbert} introduced a model named TopicBert that uses the Sentence-BERT method \citep{reimers2019sentence} for creating the graphs. These graphs are stored in a memory. Later, when other similar patterns are identified, similar graphs are categorized into a specific group, and finally, the topics are extracted. Hence, the model is combined with two parts: (1) Transformers for finding similarities, and (2) a community detection algorithm for building graphs. They also benefit from NER in order to consider the impact of various terms.

As mentioned, the main shortcoming of document-based methods is that clusters are formed only based on the co-occurrence of words without considering the impact of frequent words.
On the other hand, the reviewed feature-pivot methods are capable of identifying recurring word co-occurrence patterns and topics. However, in addition to word co-occurrences, it is required to also consider the document semantics and their connections which are missing in this group of techniques.

\subsection{Classification-based methods}
In classification-based methods, different algorithms are being used to find whether a document or text is going to represent an event or not \cite{afyouni2022multi}. \citet{ali2021traffic} proposed a method in which, first, a query-based approach is used to collect data, then by using an OLDA-based model and bidirectional long short-term memory (Bi-LSTM) each sentence is labeled individually to extract the relevant sentences for events. \citet{huang2021similarity} introduced a model which is mainly based on clustering. First, a two-step classification is utilized to dive data into two groups. Then, the cluster of events is outputted by using Bi-LSTM, expression matching, and other features related to social media texts. \citet{hettiarachchi2022embed2detect} proposed a new approach named Embed2Detect which semantic word embedding is used with hierarchical agglomerative clustering, and the combination overcomes the  limitation of previous studies.

The papers' task is to propose a novel approach for topic modeling, as a result, the models that are evaluated with our methods are only feature-based and document-based methods.

The proposed methods extract keywords from tweets to model the events in all of the mentioned methods. Hence, their task is finding main topics. In this paper, we propose a new approach for finding topics of events in social media.

The mentioned shortcomings of the previous studies motivated us to propose a model such that in addition to word co-occurrences and similar patterns, the semantic connection between the documents is also considered. The uninformative data is eliminated from the document clusters layer by layer and topics that are more closely related to the events are extracted. The proposed method tries to eliminate the demerits of the previous methods by taking advantage of their merits.

\section{Proposed Method}
\label{sec:ProposedModel}
Document-pivot methods were quite capable of identifying related documents using clustering. Furthermore, these methods were able to identify clusters of events and report the results in a fairly reasonable amount of time. The problem with these methods was the fact that different documents had entirely unrelated keywords to the topic and choosing all of the keywords in one cluster would complicate the process of identifying related words. In feature-pivot methods, this process is different. The keywords are properly identified but choosing the keywords is time-consuming. To minimize the impact of these issues, a combination of both methods must be introduced to precisely rank the clusters and choose the right keywords.

The proposed SMM method consists of 5 different modules that attempt to fix the mentioned issues using the concepts of clustering algorithms, feature-pivot methods, and their combination with a deep learning approach. The proposed method has a modular structure that eliminates unnecessary information layer by layer and can output the final result efficiently. Any stream of input data will be divided into different time intervals and the documents related to those will be processed through the five modules.

\begin{figure}[]
	\includegraphics[width= 12cm]{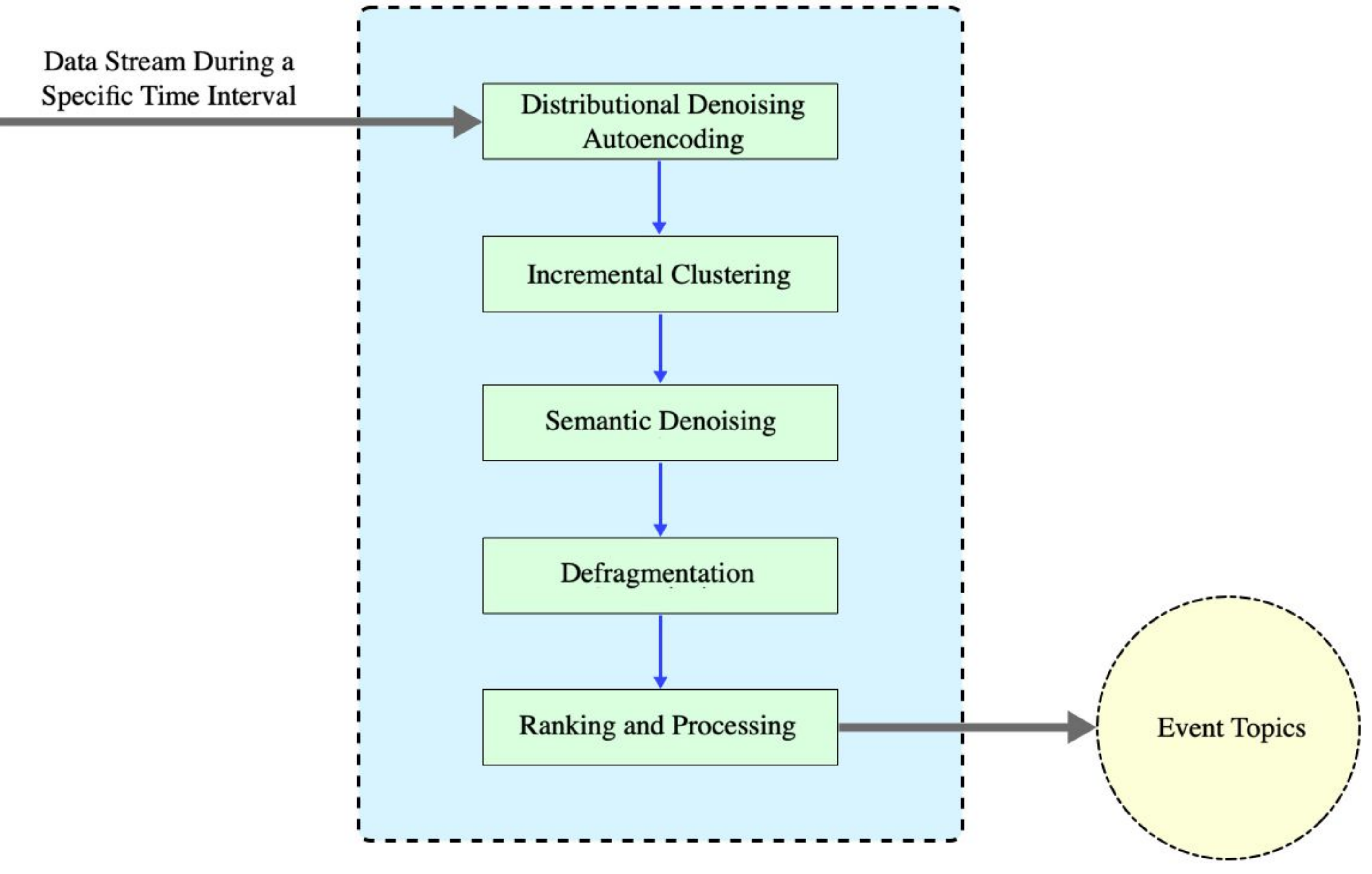}
	\centering
	\caption{Proposed Method Overview}\label{picture1}
\end{figure}

The modules of our proposed SMM method are Distributional Denoising Autoencoder, Incremental Clustering, Semantic Denoising, Defragmentation, and Ranking and Processing which are all explained in the following section. The overall structure of our proposed framework is presented in Figure \ref{picture1}.

The main contributions of our model are as follows:
\begin{itemize}
\item Ignoring unrelated tweets using deep learning has not been addressed in the previous studies. We introduced an approach to ignore unrelated tweets, improving the effectiveness and the efficiency.
\item Both semantic and world-occurrence are used in this study. The clustering algorithm does not consider semantic representation; however, we also consider the semantics of the tweets to ignore unrelated tweets.
\item An approach is introduced to address the fragmentation problem of the incremental clustering algorithm semantically.
\item A novel and efficient ranking system is proposed for events.
\end{itemize}

\subsection{Distributional Denoising Autoencoder}
When certain events regarding a football match or an election are going to be identified, people start posting documents about it on social media before the actual event takes place. For instance, in the case of a sports event, people might start posting about the winning or losing chances of teams or which player is going to score a goal. Also, in the case of an election, tweets are going to be posted about the next president of the country. Therefore, the distribution of events can be obtained prior to the event and the topics that people post about can be expected.

In identifying the events, hundreds of documents relating to the event can be found online using a suitable hashtag. However, some users might post unrelated documents with the same hashtag and that would complicate the process of identifying the events. Eliminating unrelated tweets is a useful step to identify informative clusters properly. This will improve the accuracy of the results and can also have a positive impact on run-time speed by reducing the number of documents. In other words, by identifying the distribution of the documents before the beginning of the actual event, unrelated documents can be eliminated during the process of identifying the events. Therefore, a vector representation is needed for every document.

BERT \citep{devlin2018bert} is a transformer-based model created by Google. This model receives a large number of documents and will learn the connection between the words through deep learning. This pre-trained model can be used for representing words and can be adjusted dynamically in specific fields.

The problem with the BERT model is that it takes a lot of processing for semantic similarity search. For instance, in order to calculate the pairwise similarity among 1000 sentences, 50 million computations are required. Therefore, the BERT model is not feasible for clustering. To overcome this issue, the Sentence-BERT model \citep{reimers2019sentence} was introduced to reduce the amount of processing using the Triple Network. For instance, an operation that took about 65 hours to complete with a certain hardware was reduced to just 5 seconds. The Sentence-BERT model adds a pooling layer to the BERT model and gives a fixed-size representation for the output sentences. To train the BERT model according to these alterations, a triple network was used.
In order to gain a suitable representation of the semantics of the words in a reasonable amount of time, the Sentence-BERT model was used.

Finding noisy data in various datasets has always been of importance in traditional machine learning \citep{agrawal2015survey} and deep learning \citep{kwon2019survey}. Using Autoencoders is the most popular approach among all and it has also been useful in natural language processing \citep{baynazarov2019binary}.

\begin{figure}[!ht]
	\includegraphics[scale=0.4]{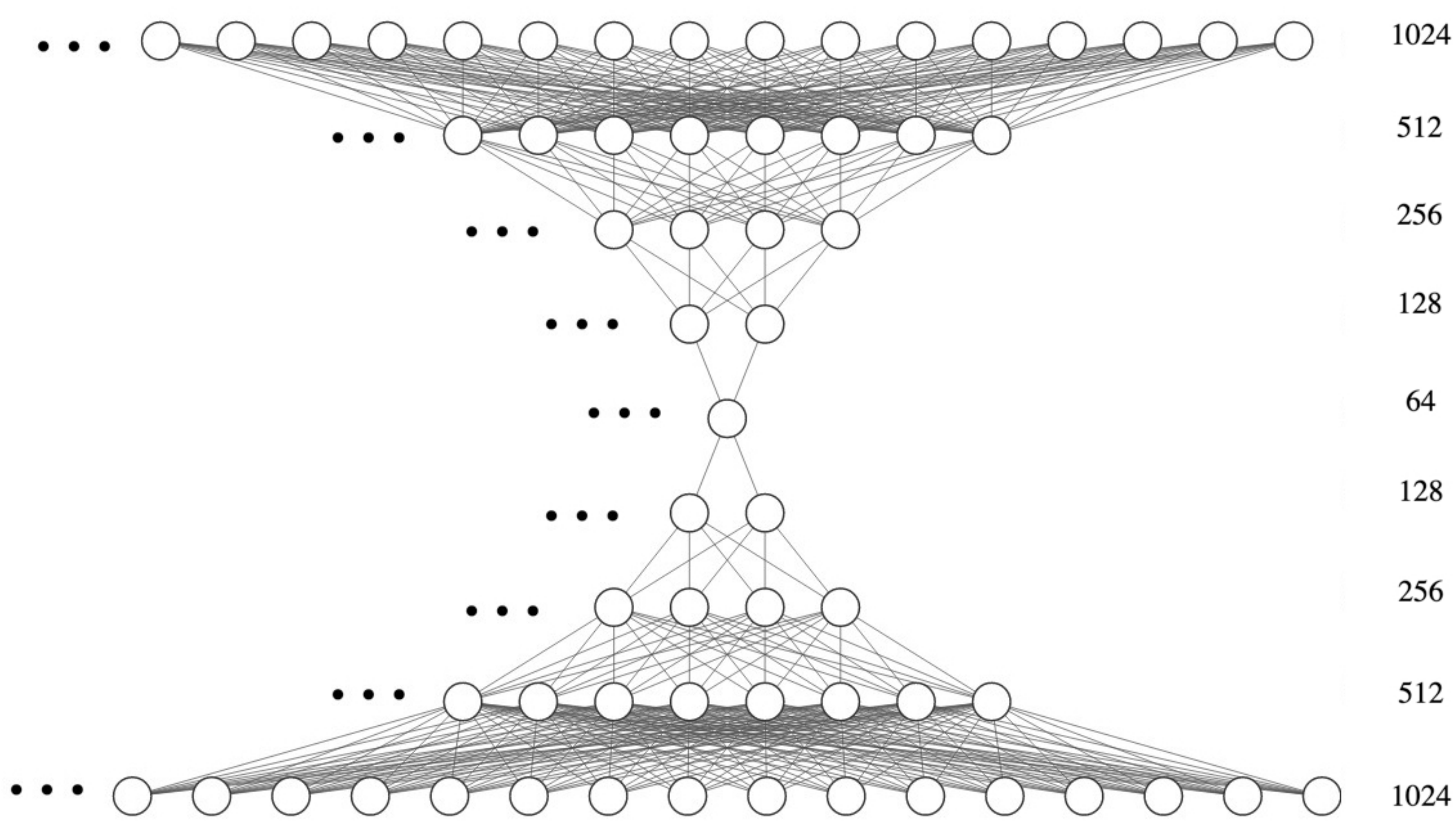}
	\centering
	\caption{Distributional Denoising Autoencoder Model Structure}\label{picture2}
\end{figure}

To obtain the distribution of the data before the actual event, the output documents of the Sentence-BERT model are used to train the autoencoder network. The input and output of the autoencoder network are vectors with the size of 1024. The structure of the autoencoder network, which is a multi-layer perceptron model, is presented in Figure \ref{picture2}.
By obtaining the distribution of the documents before the start of events using the autoencoder, outliers should be omitted using an error function which is calculated in Equation \ref{error}.

\begin{equation}
\label{error}
\mathrm{error}=\sum_{i=1}^{n}\left(Y_{i}-\hat{Y}_{i}\right)^{2}
\end{equation}

\noindent where $Y_{i}$ is the input vector and $\hat{Y}_{i}$ is the output vector of the model. This is calculated for all of the data in the time intervals. The data is then sorted and items with an error higher than ${\theta_{DDA}}$ \% of the whole data are then eliminated. Following this process, the new data is then given to the other model.

\subsection{Incremental Clustering}
In document-pivot methods, clustering algorithms are used. For instance, in Doc-p \citep{petrovic2010streaming}, and Twitternews+ \citep{hasan2019real} methods, the incremental clustering approach is used based on the TF-IDF of the words for clustering. The same approach is followed in this module, and the TF-IDF score of each word is utilized.
Based on a comparative study by \citep{mazoyer2020french}, using TF-IDF  in clustering for event detection achieved better results compared to other respresentations.

Firstly, the set of tweets posted in a particular time interval which are the output of the previous algorithm are used to obtain TF-IDF. Then the documents’ TF-IDF representations are defined, and these vectors are used in clustering the data. In the next step, an incremental clustering algorithm proposed by \citet{repp2018extracting} is used.

\subsection{Semantic Denoising}
Each cluster consists of a set of documents that might include words unrelated to the concept of the cluster. This module presents a method in order to eliminate such unrelated information.

The incremental clustering module co-relates each document to a specific cluster with the help of the TF-IDF representation which shows the co-occurrence of words. Although this could be very useful in identifying the events, the disadvantage of this method is that the words that lack a co-occurrence would be ignored and each cluster contains documents that are not related to each other regarding the variety of the related words and the meaning of the sentences. This affects the larger clusters more. Furthermore, this problem worsens when the larger clusters have higher priorities in identifying events.

To overcome this problem, this module is dedicated to semantically denoise clusters using the Sentence-BERT model \citep{reimers2019sentence}. The process starts with calculating the representation vector for each identified cluster and then the clusters are pruned using Algorithm \ref{algorithm1}.

\begin{algorithm}
	\caption{Cluster Pruning}	\label{algorithm1}

\begin{algorithmic}
	\Require\\
	$ C \leftarrow clusters$ \\
		SentBERT $\leftarrow$ Sentence-BERT Representation of Documents in Clusters \\
		$\theta_{SD} \leftarrow Threshold$
	
	\Ensure {$L \leftarrow$ Pruned Clusters}
	
	\For{$i$ in $|C|$}
		\State  {$\mu_i$  = $0$}
		\For {$j$ in $C_i$}
          \State	{$\mu_i$ = $\mu_i$ + $SentBERT_{ij}$}
		\EndFor
		\State { $\mu_i$ = $1/m \times \mu_i$}
		\For{$j$ in $C_i$}
			\If {Cosine($\mu_{ij}$, $SentBERT_{ij}$) > $\theta_{SD}$ }
			    \State {$L_{i,j} = C_{i,j}$}
		    \EndIf
		\EndFor
		\If{$C_i < 3$ }
		    \State	{$L_i$ = $\emptyset$}
		\EndIf
	\EndFor
	\State \Return $L$
%\KwRet{L}
	
\end{algorithmic}

\end{algorithm}

This question might be raised that in the case of a small number of documents in the clusters, pruning might not make any sense. To address this issue, in the ranking module, a method is used to dismiss unrelated patterns to the events.
In this module, the semantics of the documents are considered as well as the co-occurrences in order to prevent the noisy data from entering the next step.

\subsection{Defragmentation}
Similar to Twitternews+ \cite{hasan2019real}, small clusters are formed in incremental clustering that are semantically close to the bigger clusters. These small clusters cause two problems in the model. The first problem is that the smaller branches are overlooked and are pruned at the end. The other problem worth mentioning is that related small clusters are similar to large clusters; this decreases the importance of large clusters and causes them to achieve a lower rank.

Defragmentation is solved using the K-means algorithm for clustering the cluster centers. Therefore, similar clusters which illustrate specific events can be merged into one cluster. The steps can be seen in Algorithm \ref{algorithm2}.

\begin{algorithm}
	\caption{Defragmentation}	\label{algorithm2}

\begin{algorithmic}
	\Require{\\
	$L \leftarrow$  Pruned Clusters \\
	SentBERT $\leftarrow$ Sentence-BERT Representation of Documents in Pruned Clusters \\
	$K_D \leftarrow$  Number of Clusters}
	\Ensure{$D \leftarrow$ Defragmented Clusters}
	
	\For{$i$ in $|L|$}
		\State{$\mu_i$  = $0$}
		\For{$j$ in $C_i$}
			\State{$\mu_i$ = $\mu_i$ + $SentBERT_{ij}$}
		\EndFor
		\State{$\mu_i$ = $1/m \times \mu_i$}
	\EndFor
	\State{$\lambda$ = $K-means(\mu, K_D)$}
	\State{$D$ $\leftarrow$ Congregate Clusters in $C$ if Their Centers are in the Same Cluster as $\lambda$}
	\State \Return $D$
	%\KwRet{$D$}
	
\end{algorithmic}

\end{algorithm}

The difference between the approach taken in this section and the defragmentation method in the Twitternews+ framework is that Twitternews+ performs the defragmentation of clusters during the process of clustering, whereas in our proposed method, defragmentation is performed after clustering. Because all of the documents and clusters are collected throughout a specific time interval and there is no need for it to be incremental and simultaneous with clustering.

\subsection{Ranking and Processing}
The output from the previous modules was a processed model with a minimum amount of outliers. However, a mechanism has not yet been introduced for ranking, processing, and extracting the keywords from the clusters. This module aims at solving this problem through the following steps:

\begin{itemize}
	\item {Ranking: Larger clusters have a higher chance of introducing a more important event. Now consider a situation where there are unrelated tweets to the topic that are duplicated or have quoted a duplicated tweet. In this case, there might be a cluster consisting of 4 identical tweets that do not show a related event. Therefore, not only the size of the clusters must be taken into consideration, but also the number of repetitions for words in each cluster. Therefore, a combination of both factors must be used for ranking. To this aim, the following metrics are introduced for ranking the clusters that can be seen in Equations \ref{eq1} and \ref{eq2}.
	
	\begin{equation}
	\label{eq1}
	{score}_{word_{n}}=\frac{1}{m} \sum_{j=1}^{docs} \sum_{i=1}^{words}{score}_{i j}
	\end{equation}
	
	\begin{equation}
	\label{eq2}
	{score}_{{n}}=\log(score_{words_{n}}) \times \log(count_{cluster_{n}})
	\end{equation}
	
	\noindent where ${score_{i j}}$ is the number of repetitions for the word $i$ in cluster $j$ in the whole time interval, $m$ is the number of words in the cluster, and $count_{cluster_{n}}$ is the number of documents available in the cluster. Finally, ${score}_{{n}}$ reveals the score of each cluster, and then the clusters are ranked accordingly.
}

\item { Elimination of infrequent words: As can be seen in the definition of identifying an event, each set of words that are chosen as the topic of an event must have been repeated for a specific number of times. Unigrams are also very important in events and identifying frequent unigrams is only possible through their repetition in a specific interval of time. By using this idea, various words have been found in the text and are sorted according to their repetition number. Then, the keywords that have been repeated more than $\theta_{RP}$  \% of the other keywords are chosen and the rest would not be considered anymore.
	
}
\item { Elimination of clusters with fewer words: Each cluster must have a number of at least ${count_{RP}}$ keywords.
}
\item { Choosing keywords in cluster: Larger clusters have a higher rank and have more keywords. Clusters with higher ranks might place various topics in one cluster, even though these topics are repeated in lower-ranked clusters. To obtain more useful keywords and identify the main topic of the cluster, keywords are sorted according to their number of repetitions in a time interval in the cluster, and then the number of chosen keywords for each cluster is calculated according to Equation \ref{eq3}.
	
	\begin{equation}
	\label{eq3}
	count_{n}=\beta_1 + \beta_2 \times [{\frac {n}{\beta_3}}]
	\end{equation}
	
	\noindent where $\beta_1$, $\beta_2$ and $\beta_3$ are adjustable parameters to improve the accuracy, and $n$ is the rank of the cluster among others. It is visible in the equation that by increasing the depth, the number of considered keywords also increases.
}
\end{itemize}

Finally, a set of events with various topics that each consist of different keywords is outputted.

\section{Results}
\label{sec:Results}

\subsection{Datasets}
To evaluate the proposed method and comparing it to previous methods, three datasets were used which are described in the following \citep{aiello2013sensing}:

\begin{enumerate}
	\item FACup: FAC football match is the most popular match among the fans of this sport. This dataset is gathered from the 2012 final match where Chelsea beat Liverpool 2-1. Three goals were scored in this match. The events were examined throughout the 90 minutes of the match and the 15-minute break, and according to the news reports, 12 topics were considered for the events.
	
	\item SuperTuesday: In the American election system, a number of people are nominated from each party. An election is held in various states to choose one candidate to represent each party for the main election. This election starts in January, and it takes up to June. Each state holds this election on a specific day. Some states hold it on the first Tuesday of March and is considered to be an important event. The tweets regarding this event were collected in this dataset and 22 topics were chosen.
	
	\item USElection: This dataset belongs to the 6th of July 2014 presidential election in the USA where Barack Obama was elected as the president and Joe Biden was appointed as the vice president. 64 topics were identified and considered as golden data.
\end{enumerate}

The mentioned datasets have been used for proposing different event detection and topic modeling for several years, and recent studies are based on the datasets.

Considering that the task is topic modeling, for every time step, there would be keywords that represent events. An example from the FACup dataset is illustrated in Table \ref{tableer}.

\begin{table}[h]
	\begin{center}
	\begin{minipage}{\textwidth}
	\caption{{Examples from FACup Dataset}}
	\label{tableer}
	\centering

	\begin{tabular}{p{1.5cm} p{4.5cm} p{5cm}}
		\toprule
		Time Step & Golden Keywords & Relevant Tweet \\
		\midrule
		16:26 & ramires chelsea goal 1-0 1 0 score & GOAL!! Chelsea 1 - 0 Liverpool. \#FACup \\
		\midrule

		17:24 & goal  2-0 2 0 didier drogba chelsea score & Excellent goal. I think that's game over. Drogba. Chelsea 2 Liverpool 0. \#lfc \#cfc \\
		\hline
	\end{tabular}
	\end{minipage}
	\end{center}
	\end{table}

\begin{table}[h]
	\begin{center}
\begin{minipage}{\textwidth}
	\caption{Tested Dataset Information}
	\label{table1}
	\renewcommand{\arraystretch}{2}
	\makebox[\linewidth]{
	\begin{tabular}{@{}llll@{}}
		\toprule
		Dataset & FACup & SuperTuesday & USElection \\
		%\hline
		\midrule
		Data Collection Starting Time & 14:00 May 5th & 17:00 March 6th & 17:00 November 6th \\
		%\hline

		Time of the First Event & 16:16 May 5th & 22:00 March 6th & 00:00 November 7th \\
		%\hline

		Time of the Last Event & 18:10 May 5th & 08:00 March 7th & 06:50 November 7th \\
		%\hline

		Data Collection Ending Time & 20:00 May 5th & 17:00 March 7th & 05:00 November 8th \\
		%\hline

		Number of Topics & 12 & 22 & 64\\
		%\hline

		Number of Tweets & 181882 & 456129 & 1906097\\
		\hline

	\end{tabular}
}
	\end{minipage}
\end{center}
\end{table}

The statistics and information of the three datasets are presented in Table \ref{table1}.
As can be seen, the FACup dataset varies from the other datasets in terms of word distribution and tweet structure. This dataset has a lower variety of words, less complicated sentences, and also a lower number of tweets, which these features contribute to simplifying the process of identifying the events.

\subsection{Evaluation Metrics}
In the gold datasets, we have different time steps, and different keywords represent the events in each time step. As a result, the main goal of this study is finding related keywords and ignoring unrelated ones to achieve the best results.

To evaluate the proposed method, we use the following metrics that are widely used in evaluating the majority of the algorithms and models introduced in Section 2. The evaluation metrics used to compare the proposed method with past methods are based on recent studies\citep{saeed2019enhanced,asgari2020topicbert}.

\begin{enumerate}
	\item Topic-Recall: This metric is the ratio of the number of golden topics that were correctly identified among the top K topics to the total number of golden topics. Each golden topic consists of a set of keywords that are either mandatory, optional, or forbidden. A topic is gold when it includes the mandatory keywords but not the forbidden keywords.
	\item Keyword-Precision: This metric is the ratio of the total number of correctly identified keywords to the total number of identified keywords. To measure this metric, all of the mandatory and optional keywords must be calculated.
\end{enumerate}

\subsection{Preprocessing}
For each tweet in the three datasets, the following steps are followed for the preprocessing step:
\begin{enumerate}
	\item Removing words containing \# and @ from every tweet
	\item Reducing every tweet to its root by stemming
	\item Removing emojis, URLs, and stop words
	\item Removing tweets with less than two words (Leaving out the \#, @, and stop words)
	\item Removing special characters (Such as \$, \%, and etc.)
\end{enumerate}

\subsection{Hyperparameters}
In the proposed method, a set of hyperparameters are required. The set of parameters and their values are presented in Table \ref{table2}.

Due to the high similarity of SuperTuesday and USElection datasets we use the same parameters for these two datasets. For FACup, however, we use different parameters due to its different structure, which can also be seen in other studies as well\citep{saeed2019enhanced,asgari2020topicbert}.

\begin{table}[h]
	\begin{center}
	\begin{minipage}{\textwidth}
	\caption{Hyperparameters Used in the Proposed Method for Different Datasets}
	\label{table2}
	\centering

	\begin{tabular}{@{}cccc@{}}
		\toprule
		& FACup & SuperTuesday & USElection \\
		\midrule

		%\hline
		${\theta_{DDA}}$ & 98 & 98 & 98 \\
		%\hline
		${\theta_{IC}}$ & 70 & 95 & 95 \\
		%\hline
		${IC}$ & 64 & 64 & 64 \\
		%\hline
		$\theta_{SD}$ & 85 & 85 & 85 \\
		%\hline
		$K_{D}$ & 16 & 100 & 100 \\
		%\hline
		${\theta_{RP}}$ & 80 & 80 & 80 \\
		%\hline
		$count_{RP}$ & 0 & 1 & 1 \\
		%\hline
		$\beta_1$ & 3 & 3 & 3 \\
		%\hline
		$\beta_2$ & 25 & 25 & 25 \\
		%\hline
		$\beta_3$ & 3 & 3 & 3 \\
		\hline
		%\hline
	\end{tabular}
	\end{minipage}
	\end{center}
\end{table}

\subsection{Results and Discussion}
In this section, the results are reported for each of the mentioned metrics and finally the average results are presented to better compare the methods.
We evaluate our models with the topic-recall and keyword-precision metrics. For every metric, the results for FACup, SuperTuesday, and USElection datasets are calculated. In the end, the average results of the three datasets are available for the two mentioned metrics.

In a clustering approach, there could be different clusters that represent an event, and there should be a criterion in sorting clusters and extracting keywords to get the most relevant events in higher ranks. The results are then calculated at every rank. As a result, it could be important that the model will find relevant events by matching the keywords, and for the topic-recall metric, the evaluation is based on rank. In addition to the result of each rank, the system's overall performance is also noticeable, and we include the average metrics' results for all ranks for every dataset.

In addition, considering that we combined five modules, we need to study whether the modules are effective or not. Hence, we also show the results by omitting some of the modules in order to show their impact.

\subsubsection{Topic-Recall Evaluation}
By examining the documents according to their ranks, the effectiveness of the algorithms and their impact on identifying the events are concluded.

The results for the topic-recall metric of the FACup dataset can be seen in  Table \ref{table3}. According to the results, the highest precision belongs to the TopicBERT model, which is approximately 4\% more precise than the proposed method. Both models are able to identify all of the topics from rank 8. It is concluded that the TopicBERT model can have better results for smaller datasets in this metric.

\begin{table}[]
	%\footnotesize
	\centering
	\begin{minipage}{\textwidth}
	\caption{Topic-Recall Metric Evaluation Results for the FACup Dataset}
	\label{table3}
	\renewcommand{\arraystretch}{1.5}
	\makebox[\linewidth]{
		\begin{tabular}{@{}p{2cm}ccccccccccc@{}}
			%\hline
			\toprule
			\multirow{2}{*}{Model} &
			\multirow{2}{*}{Average} &
			\multicolumn{10}{c}{Topic Rank} \\ \cline{3-12}
			& & 2 & 4 & 6 & 8 & 10 & 12 & 14 & 16 & 18 & 20 \\
			%\hline
			\midrule
			LDA\cite{blei2003latent} & $0.817$ & $0.692$ & $0.692$ & $0.84$ & $0.84$ & $0.92$ & $0.92$ & $0.84$ & $0.84$ & $0.84$ & $0.75$ \\
			%\hline
			Doc-p\cite{petrovic2010streaming} & $0.946$ & $0.769$ & $0.85$ & $0.92$ & $0.92$ & \bfseries{1} & \bfseries{1} & \bfseries{1} & \bfseries{1} & \bfseries{1} & \bfseries{1} \\
			%\hline
			Gfeat-p\cite{o2010tweetmotif} & $0.324$ & 0 & $0.308$ & $0.308$ & $0.375$ & $0.375$ & $0.375$ & $0.375$ & $0.375$ & $0.375$ & $0.375$ \\
			%\hline
			SFPM\cite{aiello2013sensing} & $0.929$ & $0.615$ & $0.84$ & $0.84$ & \bfseries{1} & \bfseries{1} & \bfseries{1} & \bfseries{1} & \bfseries{1} & \bfseries{1} & \bfseries{1} \\
			%\hline
			BNGram\cite{aiello2013sensing} & $0.905$ & $0.769$ & $0.92$ & $0.92$ & $0.92$ & $0.92$ & $0.92$ & $0.92$ & $0.92$ & $0.92$ & $0.92$ \\
			%\hline
			SVD+Kmeans\cite{nur2015combination} & $0.828$ & $0.482$ & $0.569$ & $0.71$ & $0.824$ & $0.938$ & $0.951$ & $0.951$ & $0.951$ & $0.951$ & $0.951$ \\
			%\hline
			SNMF-Orig\cite{prabandari2017comparative} & $0.320$ & $0.1$ & $0.177$ & $0.254$ & $0.331$ & $0.389$ & $0.389$ & $0.389$ & $0.389$ & $0.389$ & $0.389$ \\
			%\hline
			SNMF-Kl\cite{prabandari2017comparative} & $0.681$ & $0.167$ & $0.334$ & $0.502$ & $0.67$ & $0.837$ & $0.837$ & $0.84$ & $0.85$ & $0.85$ & $0.924$ \\
			%\hline
			Exemplar\cite{elbagoury2015exemplar} & $0.894$ & $0.81$ & $0.838$ & $0.886$ & $0.908$ & $0.916$ & $0.916$ & $0.916$ & $0.916$ & $0.916$ & $0.916$ \\
			%\hline
			EHG\cite{saeed2019enhanced} & $0.815$ & $0.379$ & $0.591$ & $0.727$ & $0.727$ & $0.864$ & $0.864$ & \bfseries{1} & \bfseries{1} & \bfseries{1} & \bfseries{1} \\
			%\hline
			TopicBERT\cite{asgari2020topicbert} & \bfseries{0.971} & \bfseries{0.81} & \bfseries{0.951} & \bfseries{0.951} & \bfseries{1} & \bfseries{1} & \bfseries{1} & \bfseries{1} & \bfseries{1} & \bfseries{1} & \bfseries{1} \\
			%\hline
			SMM &
			$0.931$ & $0.538$ & $0.923$ & $0.923$ & $0.923$ & \bfseries{1} & \bfseries{1} & \bfseries{1} & \bfseries{1} & \bfseries{1} & \bfseries{1} \\
			\hline
		\end{tabular}
		
	}
	\end{minipage}
\end{table}
The results for the topic-recall metric on the SuperTuesday dataset can be seen in  Table \ref{table4}. According to the results, this model was able to show an average of 2.1\% improvement in comparison to the best case of the previous model. An improvement of over 17.3\% is visible in ranks lower than 60 which suggests the efficiency of this model for lower ranks. The same conclusion can be made for the Doc-p model that uses clustering. Assuming that the Doc-p module is approximately equal to the incremental clustering module, it can be concluded that a combination of the defragmentation and semantic denoising modules can improve the effectiveness of the clustering process.

\begin{sidewaystable}
%\sidewaystablefn

	%\footnotesize
	\begin{minipage}{\textwidth}
	\centering
	\caption{Topic-Recall Evaluation of the SuperTuesday Dataset}
	\label{table4}
	\renewcommand{\arraystretch}{1.5}
	\makebox[\linewidth]{
		\begin{tabular}{@{}p{2cm}cccccccccccc@{}}
			%\hline
			\toprule
			\multirow{2}{*}{Model} &
			\multirow{2}{*}{Average} &
			\multicolumn{11}{c}{Topic Rank} \\ \cline{3-13}
			& & 2 & 10 & 20 & 30 & 40 & 50 & 60 & 70 & 80 & 90 & 100 \\
			\midrule
			%\hline
			LDA\cite{blei2003latent} & $0.161$ & 0 & 0 & 0 & $0.18$ & $0.13$ & $0.13$ & $0.18$ & $0.28$ & $0.28$ & $0.37$ & $0.227$ \\
			%\hline
			Doc-p\cite{petrovic2010streaming} & $0.457$ & $0.227$ & $0.227$ & $0.31$ & $0.4$ & $0.46$ & $0.5$ & $0.5$ & $0.5$ & $0.54$ & $0.68$ & $0.68$ \\
			%\hline
			Gfeat-p\cite{o2010tweetmotif} & $0.206$ & $0.046$ & $0.045$ & $0.085$ & $0.18$ & $0.227$ & $0.28$ & $0.28$ & $0.28$ & $0.28$ & $0.28$ & $0.28$ \\
			%\hline
			SFPM\cite{aiello2013sensing} & $0.294$ & $0.182$ & $0.182$ & $0.27$ & $0.325$ & $0.325$ & $0.325$ & $0.325$ & $0.325$ & $0.325$ & $0.325$ & $0.325$ \\
			%\hline
			BNGram\cite{aiello2013sensing} & $0.533$ & $0.5$ & $0.5$ & $0.54$ & $0.54$ & $0.54$ & $0.54$ & $0.54$ & $0.54$ & $0.54$ & $0.54$ & $0.54$ \\
			%\hline
			SVD+Kmeans\cite{nur2015combination} & $0.521$ & $0.192$ & $0.236$ & $0.4$ & $0.488$ & $0.547$ & $0.58$ & $0.626$ & $0.666$ & $0.666$ & $0.666$ & $0.666$ \\
			%\hline
			SNMF-Orig\cite{prabandari2017comparative} & $0.224$ & 0 & $0.045$ & $0.1$ & $0.183$ & $0.227$ & $0.227$ & $0.227$ & $0.32$ & $0.32$ & $0.363$ & $0.453$ \\
			%\hline
			SNMF-Kl\cite{prabandari2017comparative} & $0.290$ & 0 & $0.1$ & $0.183$ & $0.183$ & $0.318$ & $0.41$ & $0.366$ & $0.41$ & $0.453$ & $0.363$ & $0.41$ \\
			%\hline
			Exemplar\cite{elbagoury2015exemplar} & $0.557$ & $0.246$ & \bfseries{0.463} & \bfseries{0.538} & $0.572$ & $0.586$ & $0.597$ & $0.6$ & $0.617$ & $0.638$ & $0.638$ & $0.638$ \\
			%\hline
			EHG\cite{saeed2019enhanced} & $0.579$ & $0.163$ & $0.408$ & $0.466$ & $0.54$ & $0.628$ & $0.674$ & $0.674$ & $0.699$ & $0.699$ & $0.711$ & $0.711$ \\
			%\hline
			TopicBERT\cite{asgari2020topicbert} & $0.639$ & \bfseries{0.463} & $0.466$ & $0.58$ & \bfseries{0.646} & \bfseries{0.699} & 0.699 & 0.699 & $0.711$ & $0.722$ & $0.722$ & $0.722$ \\
			%\hline
			SMM &
			\bfseries{0.660} & $0.158$ & $0.211$ & $0.368$ & $0.579$ & $0.684$ & \bfseries{0.789} & \bfseries{0.895} & \bfseries{0.895} & \bfseries{0.895} & \bfseries{0.895} & \bfseries{0.895} \\
			%\botrule
			\hline
		\end{tabular}
	}
	\end{minipage}%{\textwidth}
\end{sidewaystable}

The results for the topic-recall metric on the USElection dataset can be seen in  Table \ref{table5}. According to the data displayed in this table, the TopicBERT model has an approximately 1.7\% higher accuracy than the proposed method. The proposed method has achieved higher accuracy in lower ranks.

\begin{sidewaystable}
%\sidewaystablefn
\begin{minipage}{\textwidth}
%\footnotesize
	\centering
	\caption{Topic-Recall Metric Evaluation of the USElection Dataset}
	\label{table5}
	\renewcommand{\arraystretch}{1.5}
	\makebox[\linewidth]{
			\begin{tabular}{@{}p{2cm}cccccccccccc@{}}
			%\hline
			\toprule
			\multirow{2}{*}{Model} &
			\multirow{2}{*}{Average} &
			\multicolumn{11}{c}{Topic Rank} \\ \cline{3-13}
			& & 2 & 10 & 20 & 30 & 40 & 50 & 60 & 70 & 80 & 90 & 100 \\
			\midrule
			LDA\cite{blei2003latent} & $0.110$ & $0.109$ & $0.109$ & $0.185$ & $0.245$ & $0.22$ & $0.28$ & $0.325$ & $0.5$ & $0.475$ & $0.43$ & $0.46$ \\
			%\hline
			Doc-p\cite{petrovic2010streaming} & $0.547$ & $0.234$ & $0.234$ & $0.415$ & $0.505$ & $0.56$ & $0.615$ & $0.615$ & $0.69$ & $0.69$ & $0.72$ & $0.74$ \\
			%\hline
			Gfeat-p\cite{o2010tweetmotif} & $0.158$ & $0.078$ & $0.078$ & $0.14$ & $0.18$ & $0.18$ & $0.18$ & $0.18$ & $0.18$ & $0.18$ & $0.18$ & $0.18$ \\
			%\hline
			SFPM\cite{aiello2013sensing} & $0.499$ & \bfseries{0.359} & $0.359$ & $0.465$ & $0.525$ & $0.54$ & $0.54$ & $0.54$ & $0.54$ & $0.54$ & $0.54$ & $0.54$ \\
			%\hline
			BNGram\cite{aiello2013sensing} & $0.492$ & $0.48$ & $0.48$ & $0.495$ & $0.495$ & $0.495$ & $0.495$ & $0.495$ & $0.495$ & $0.495$ & $0.495$ & $0.495$ \\
			%\hline
			SVD+Kmeans\cite{nur2015combination} & $0.545$ & $0.11$ & $0.216$ & $0.42$ & $0.522$ & $0.588$ & $0.608$ & $0.647$ & $0.7$ & $0.72$ & $0.72$ & $0.74$ \\
			%\hline
			SNMF-Orig\cite{prabandari2017comparative} & $0.383$ & $0.075$ & $0.075$ & $0.154$ & $0.218$ & $0.439$ & $0.467$ & $0.483$ & $0.545$ & $0.563$ & $0.595$ & $0.595$ \\
			%\hline
			SNMF-Kl\cite{prabandari2017comparative} & $0.474$ & $0.154$ & $0.154$ & $0.326$ & $0.4$ & $0.547$ & $0.581$ & $0.562$ & $0.618$ & $0.6$ & $0.652$ & $0.622$ \\
			%\hline
			Exemplar\cite{elbagoury2015exemplar} & $0.451$ & $0.022$ & $0.142$ & $0.244$ & $0.364$ & $0.465$ & $0.532$ & $0.59$ & $0.628$ & $0.651$ & $0.662$ & $0.662$ \\
			%\hline
			EHG\cite{saeed2019enhanced} & $0.694$ & $0.279$ & $0.608$ & \bfseries{0.67} & $0.688$ & $0.733$ & $0.746$ & $0.762$ & $0.772$ & $0.78$ & $0.796$ & $0.805$ \\
			%\hline
			TopicBERT\cite{asgari2020topicbert} & \bfseries{0.717} & $0.326$ & \bfseries{0.628} & $0.628$ & $0.733$ & $0.772$ & $0.772$ & $0.78$ & $0.805$ & $0.805$ & $0.813$ & $0.821$ \\
			%\hline
			SMM &
			$0.700$ & $0.031$ & $0.5$ & $0.656$ & \bfseries{0.734} & \bfseries{0.781} & \bfseries{0.797} & \bfseries{0.828} & \bfseries{0.844} & \bfseries{0.844} & \bfseries{0.844} & \bfseries{0.844} \\
			\hline
			%\botrule
		\end{tabular}
	}
	\end{minipage}
\end{sidewaystable}

Overall, based on the results, for the lower ranks, the proposed model is undoubtedly the most accurate. In higher ranks, however, the TopicBERT model can be a serious contender. In other words, the proposed method can find more topics than other methods in the first 100 extracted events.

As mentioned,  in our proposed framework two modules, namely the distributional denoising autoencoder module and the ranking and processing module, provide the main contributions of this study. Therefore, in the next step of our experiment, we study the impact of these two modules individually. To this aim, the results for the proposed method without the distributional denoising autoencoder module and the ranking and processing module on the FACup, SuperTuesday, and USElection datasets are reported in Tables \ref{table6}, \ref{table7}, and  \ref{table8}, respectively.

As can be seen, the proposed method is 2.3\% more efficient after eliminating the ranking and processing module. The reason behind this is that the respective model has eliminated the main keywords due to the small and limited size of this dataset.
%ِOn the SuperTuesday datasets, however, due to the high complexity of the tweets in this dataset, the ranking and processing module performed well and was able to improve the results by 12.4\%.
The results of the proposed method on the USElection dataset without the distributional denoising autoencoder are somehow equal to the results of the proposed method without the ranking and processing module and are both less accurate than the proposed method by 2\%,  due to the high complexity of tweets in this dataset.

According to the obtained results, overall, we observe that both the distributional denoising autoencoder module and the ranking and processing module improve the results.

In general, the difference between the proposed method and past studies is that the model is more effective in finding a substantial number of topics in low ranks in both small and big datasets, which can be helpful for finding all related events.

\begin{table}[h]
	%\footnotesize
	\begin{minipage}{\textwidth}
	\centering
	\caption{The Results of the Elimination of Different Modules on the FACup Dataset}
	\label{table6}
	\renewcommand{\arraystretch}{1.5}
	\makebox[\linewidth]{
		\begin{tabular}{@{}p{2.4cm}ccccccccccc@{}}
			%\hline
			\toprule
			\multirow{2}{*}{Model} &
			\multirow{2}{*}{Average} &
			\multicolumn{10}{c}{Topic Rank} \\ \cline{3-12}
			& & 2 & 4 & 6 & 8 & 10 & 12 & 14 & 16 & 18 & 20 \\
			\midrule
			SMM without DDA &
			$0.861$ & $0.538$ & $0.846$ & $0.846$ & $0.846$ & $0.923$ & $0.923$ & $0.923$ & $0.923$ & $0.923$ & $0.923$ \\
			%\hline
			SMM without RP &
			\bfseries{0.954} & \bfseries{0.769} & $0.846$ & \bfseries{0.923} & \bfseries{1} & \bfseries{1} & \bfseries{1} & \bfseries{1} & \bfseries{1} & \bfseries{1} & \bfseries{1} \\
			%\hline
			SMM &
			$0.931$ & $0.538$ & \bfseries{0.923} & \bfseries{0.923} & $0.923$ & \bfseries{1} & \bfseries{1} & \bfseries{1} & \bfseries{1} & \bfseries{1} & \bfseries{1} \\
			\hline
		\end{tabular}
	}
	\end{minipage}
\end{table}

\begin{sidewaystable}
%\sidewaystablefn
\begin{minipage}{\textwidth}
	\centering
	
	\caption{The Results of the Elimination of Different Modules on the SuperTuesday Dataset}
	\label{table7}
	\renewcommand{\arraystretch}{1.5}
	\makebox[\linewidth]{
		\begin{tabular}{@{}p{2.6cm}cccccccccccc@{}}
			%\hline
			\toprule
			\multirow{2}{*}{Model} &
			\multirow{2}{*}{Average} &
			\multicolumn{11}{c}{Topic Rank} \\ \cline{3-13}
			& & 2 & 10 & 20 & 30 & 40 & 50 & 60 & 70 & 80 & 90 & 100 \\
			\midrule
			SMM without DDA &
			$0.531$ & $0.053$ & $0.263$ & \bfseries{0.421} & $0.421$ & $0.526$ & $0.579$ & $0.632$ & $0.737$ & $0.737$ & $0.737$ & $0.737$ \\
			%\hline
			SMM without RP &
			$0.536$ & $0.105$ & \bfseries{0.316} & \bfseries{0.421} & $0.421$ & $0.526$ & $0.579$ & $0.632$ & $0.684$ & $0.737$ & $0.737$ & $0.737$ \\
			%\hline
			SMM &
			\bfseries{0.660} & \bfseries{0.158} & $0.211$ & $0.368$ & \bfseries{0.579} & \bfseries{0.684} & \bfseries{0.789} & \bfseries{0.895} & \bfseries{0.895} & \bfseries{0.895} & \bfseries{0.895} & \bfseries{0.895} \\
			\hline
			%\botrule
		\end{tabular}
	}
	\end{minipage}
\end{sidewaystable}

\begin{sidewaystable}

	%\footnotesize
	\centering
	\begin{minipage}{\textwidth}
	\caption{The Results of the Elimination of Different Modules on the USElection Dataset}
	\label{table8}
	\renewcommand{\arraystretch}{1.5}
	\makebox[\linewidth]{
		\begin{tabular}{@{}p{2.6cm}cccccccccccc@{}}
			%\hline
			\toprule
			\multirow{2}{*}{Model} &
			\multirow{2}{*}{Average} &
			\multicolumn{11}{c}{Topic Rank} \\ \cline{3-13}
			& & 2 & 10 & 20 & 30 & 40 & 50 & 60 & 70 & 80 & 90 & 100 \\
			\midrule
			SMM without DDA &
			$0.680$ & $0.016$ & $0.344$ & $0.594$ & $0.719$ & \bfseries{0.781} & \bfseries{0.828} & \bfseries{0.828} & \bfseries{0.844} & \bfseries{0.844} & \bfseries{0.844} & \bfseries{0.844} \\
			%\hline
			SMM without RP &
			$0.680$ & \bfseries{0.125} & $0.469$ & $0.609$ & $0.719$ & \bfseries{0.781} & $0.797$ & $0.797$ & $0.797$ & $0.797$ & $0.797$ & $0.797$ \\
			%\hline
			SMM &
			\bfseries{0.700} & $0.031$ & \bfseries{0.5} & \bfseries{0.656} & \bfseries{0.734} & \bfseries{0.781} & $0.797$ & \bfseries{0.828} & \bfseries{0.844} & \bfseries{0.844} & \bfseries{0.844} & \bfseries{0.844} \\
			\hline
			%\botrule
		\end{tabular}
	}
	\end{minipage}
\end{sidewaystable}

\vspace{15mm}

\subsubsection{Keyword-Precision Evaluation}
To calculate this metric, keywords of two top-ranked events are taken into consideration.  This helps us to identify how the keywords are connected to each other in more important topics and what percentage of them can give more useful information.

The results of the keyword-precision evaluation for the mentioned datasets is displayed in Table \ref{table9}. The proposed method is able to significantly improve the results on the USElection and FACup datasets. In the SuperTuesday dataset, however, the TopicBERT model has a better performance than the proposed method.

\begin{table}[]
	\begin{minipage}{\textwidth}
	\caption{Dataset Results for the Keyword-Precision Metric in Different Models}
	\label{table9}
	\renewcommand{\arraystretch}{1.2}
		\centering

	\begin{tabular}{@{}p{2.4cm}ccc@{}}
      \toprule
		\multirow{2}{*}{Model} & \multicolumn{3}{c}{The Topic Rank Equals 2} \\
		& FACup & SuperTuesday & USElection \\
		\midrule
		LDA\cite{blei2003latent} & $0.164$ & 0 & $0.165$ \\
		%\hline
		Doc-p\cite{petrovic2010streaming} & $0.337$ & $0.511$ & $0.401$ \\
		%\hline
		Gfeat-p\cite{o2010tweetmotif} & 0 & $0.375$ & $0.375$ \\
		%\hline
		SFPM\cite{aiello2013sensing} & $0.233$ & $0.471$ & $0.241$  \\
		%\hline
		BNGram\cite{aiello2013sensing} & $0.299$ & $0.628$ & $0.405$  \\
		%\hline
		SVD+Kmeans\cite{nur2015combination} & $0.242$ & $0.367$ & $0.3$  \\
		%\hline
		SNMF-Orig\cite{prabandari2017comparative} & $0.33$ & $0.241$ & $0.241$  \\
		%\hline
		SNMF-Kl\cite{prabandari2017comparative} & $0.242$ & $0.164$ & $0.164$  \\
		%\hline
		Exemplar\cite{elbagoury2015exemplar} &  $0.3$ & $0.485$ & $0.391$  \\
		%\hline
		EHG\cite{saeed2019enhanced} & $0.442$ & $0.812$ & $0.591$  \\
		%\hline
		TopicBERT\cite{asgari2020topicbert} & $0.456$ & \bfseries{0.851} & $0.621$  \\
		%\hline
		SMM &
		\bfseries{0.667} & $0.667$ & \bfseries{0.833}  \\
		%\hline
		\hline
	\end{tabular}
	\end{minipage}
\end{table}

The impact of eliminating different modules on the datasets is displayed in Table \ref{table10}. It can be seen that in this metric, eliminating the ranking and processing module has significantly lowered the performance which indicates the importance of this module in the proposed method. In addition, we can see that the distributional denoising autoencoder has improved the results too.

In Average, the proposed method can find more related keywords than past models in important and top rank events. In addition, the model also can be helpful in finding more related event topics in low ranks.

\begin{table}[]
	\begin{minipage}{\textwidth}
	\caption{The Results of the Elimination of Different Modules on the datasets for keyword-precision Metric}
	\label{table10}
	\renewcommand{\arraystretch}{1.2}
		\centering

	\begin{tabular}{@{}p{3cm}ccc@{}}
      \toprule
		\multirow{2}{*}{Model} & \multicolumn{3}{c}{The Topic Rank Equals 2} \\
		& FACup & SuperTuesday & USElection \\
		\midrule
		SMM without DDA &
		$0.65$ & \bfseries{0.667} & $0.667$ \\
	%\hline
		SMM without RP &
		$0.391$ & $0.139$ & $0.139$ \\
		%\hline
		SMM &
		\bfseries{0.667} & \bfseries{0.667} & \bfseries{0.833} \\
		%\hline
		\hline
	\end{tabular}
	\end{minipage}
\end{table}

\subsubsection{Evaluation based on the Average Results of the Metrics}
For a better comparison, the average results of the topic-recall and keyword-precision metrics are displayed in Table \ref{table11}. According to the results, the proposed method has shown an approximate 7.9\% improvement in the keyword-precision metric compared to the TopicBERT model and achieved relatively competitive results to the TopicBERT's performance in the topic-recall metric. It is concluded that the proposed method has a better performance for lower ranked documents in topic-recall metric and is able to find better keywords in higher ranked documents on average.

\begin{table}[]
	\begin{minipage}{\textwidth}

	\caption{Average Results of the Topic-Recall and Keyword-Precision Metrics}
	\label{table11}
	\renewcommand{\arraystretch}{1.2}
		\centering

	\begin{tabular}{@{}p{2.4cm}cc@{}}
		\toprule
		Model & Average Topic-Recall & Average Keyword-Precision \\
		\midrule
		LDA\cite{blei2003latent} & $0.363$ & $0.110$ \\
		%\hline
		Doc-p\cite{petrovic2010streaming} & $0.650$ & $0.416$ \\
		%\hline
		Gfeat-p\cite{o2010tweetmotif} & $0.229$ & $0.25$ \\
		%\hline
		SFPM\cite{aiello2013sensing} & $0.574$ & $0.315$ \\
		%\hline
		BNGram\cite{aiello2013sensing} & $0.6433$ & $0.444$ \\
		%\hline
		SVD+Kmeans\cite{nur2015combination} & $0.631$ & $0.303$ \\
	%	\hline
		SNMF-Orig\cite{prabandari2017comparative} & $0.309$ & $0.271$ \\
		%\hline
		SNMF-Kl\cite{prabandari2017comparative} & $0.482$ & $0.19$ \\
	%	\hline
		Exemplar\cite{elbagoury2015exemplar} & $0.634$ & $0.392$ \\
	%	\hline
		EHG\cite{saeed2019enhanced} & $0.696$ & $0.615$ \\
	%	\hline
		TopicBERT\cite{asgari2020topicbert} & \bfseries{0.778} & $0.643$ \\
	%	\hline
		SMM &
		$0.764$ & \bfseries{0.722} \\
	%	\hline
	\hline
	\end{tabular}
	\end{minipage}
\end{table}

The effectiveness of different modules on the datasets for the average results of the mentioned metrics is displayed in Table \ref{table12}. According to the results, the ranking and processing module increases the performance of the method by 4.1\% in the average topic-recall metric. The performance increases in the average keyword-precision metric by 49.9\%. By adding the distributional denoising autoencoder module, the performance increases by 6.1\% and 7.2\% for the keyword-precision and topic-recall metrics, respectively. The results indicate the impact of these modules which are the main contributions of our model.

\begin{table}[]
	\begin{minipage}{\textwidth}
%\centering
	\caption{The Results of the Elimination of Different Modules in Average Metrics}
	\label{table12}
	\renewcommand{\arraystretch}{1.2}
	\centering

	\begin{tabular}{@{}p{2.8cm}cc@{}}
		\toprule
		Model & Average Topic-Recall & Average Keyword-Precision \\
		\midrule
		SMM without DDA &
		$0.691$ & $0.661$ \\
		%\hline
		SMM without RP &
		$0.723$ & $0.223$ \\
		%\hline
		SMM &
		\bfseries{0.764} & \bfseries{0.722} \\
		%\hline
		\hline
	\end{tabular}
	\end{minipage}
\end{table}

\vspace{5mm}

\section{Conclusion and Future Work}
\label{sec:Conclusion}

%\subsection{Conclusion}
The growing use of social media causes billions of messages to be shared on the internet on a daily basis. A group of these documents might report a specific concept or inform us about an event. These events might happen in various time intervals or locations. Identifying these events has been widely investigated in the past few years, where many of the past research studies aimed to identify the events using Twitter.

Event detection in the literature is divided into three different methods: Document-based, Feature-based, and Classification-based methods, which have their respective limitations. This research introduces a novel method to improve the aforementioned methods using a modular structure. The proposed method consists of 5 modules, namely distributional denoising autoencoder, incremental clustering, semantic denoising, defragmentation, and ranking and processing module.

The proposed method was compared to 11 state-of-the-art methods using three datasets, FACup, SuperTuesday, and USElection. The results showed the superiority of the proposed model compared to 10 methods.  Compared to the TopicBERT model, our method showed 1.4\% lower performance in the topic-recall metric, but 7.9\% improvement in the keyword-precision metric.

%\subsection{Future Work}
In this research, the proposed method was compared to 11 other methods. Each of these methods had a number of advantages. Our modular architecture gives us the opportunity to use various algorithms in different states. Using these models as modules can result in a better and different outcome. For instance, the TopicBERT algorithm can be applied on the clusters before entering the ranking and processing module in order to extract more useful keywords.

Furthermore, the majority of these methods use static parameters. By using reinforcement learning \citep{sehgal2019deep}, these parameters can be improved through time in order to better identify the data distribution changes for different topics.

\bibliographystyle{plainnat}
\bibliography{references}

\end{document}